\documentclass[letterpaper]{article} 
\usepackage{aaai23}  
\usepackage{times}  
\usepackage{helvet}  
\usepackage{courier}  
\usepackage[hyphens]{url}  
\usepackage{graphicx} 
\usepackage{natbib}  
\usepackage{caption} 
\usepackage{algorithm}
\usepackage{algorithmic}

\usepackage{newfloat}
\usepackage{listings}
\DeclareCaptionStyle{ruled}{labelfont=normalfont,labelsep=colon,strut=off} 
\floatstyle{ruled}
\newfloat{listing}{tb}{lst}{}
\floatname{listing}{Listing}
\usepackage{amsmath}
\usepackage{amsfonts}
\usepackage{amssymb}
\usepackage{booktabs}
\usepackage{pifont}
\usepackage{mathrsfs}
\usepackage{amsthm}
\usepackage{tasks}
\usepackage{enumitem}
\usepackage{booktabs}

\providecommand{\calA}{\mathcal{A}}
\providecommand{\calC}{\mathcal{C}}
\providecommand{\calI}{\mathcal{I}}
\providecommand{\calJ}{\mathcal{J}}
\providecommand{\calM}{\mathcal{M}}
\providecommand{\calN}{\mathcal{N}}
\providecommand{\calS}{\mathcal{S}}

\providecommand{\sym}[1]{\mathcal{S}^{#1}}
\providecommand{\spd}[1]{\mathcal{S}^{#1}_{++}}

\providecommand{\bbR}[1]{\mathbb {R}^{#1}}

\providecommand{\cinf}{C^{\infty}}

\providecommand{\mor}{{\mathrm{Mor}}}
\providecommand{\obj}{{\mathrm{Obj}}}
\providecommand{\idmap}{1}

\providecommand{\secref}[1]{Section~\ref{#1}}

\providecommand{\propsref}[1]{Proposition~\ref{#1}}
\providecommand{\thmref}[1]{Theorem~\ref{#1}}

\providecommand{\tabref}[1]{Table~\ref{#1}}
\providecommand{\figref}[1]{Figure~\ref{#1}}

\newtheorem{theorem}{Theorem}

\newtheorem{props}{Proposition}

\theoremstyle{definition}
\newtheorem{definition}{Definition}

\theoremstyle{remark}

\title{Riemannian Local Mechanism for SPD Neural Networks}

\author {
    Ziheng Chen\textsuperscript{\rm 1},
    Tianyang Xu\textsuperscript{\rm 1},
    Xiao-Jun Wu\footnote{Corresponding author.}\textsuperscript{\rm 1},
    Rui Wang\textsuperscript{\rm 1},
    Zhiwu Huang\textsuperscript{\rm 2},
    Josef Kittler\textsuperscript{\rm 3}
}
\affiliations {
    \textsuperscript{\rm 1}School of Artificial Intelligence and Computer Science, Jiangnan University \\
    \textsuperscript{\rm 2}School of Computing and Information Systems, Singapore Management University\\
    \textsuperscript{\rm 3}Centre for Vision, Speech and Signal Processing (CVSSP), University of Surrey\\
    ziheng\_ch@163.com,
    \{tianyang.xu, wu\_xiaojun,cs\_wr\}@jiangnan.edu.cn,
    zwhuang@smu.edu.sg,
    j.kittler@surrey.ac.uk
}

\begin{document}
\maketitle
\begin{abstract}
The Symmetric Positive Definite (SPD) matrices have received wide attention for data representation in many scientific areas.
Although there are many different attempts to develop effective deep architectures for data processing on the Riemannian manifold of SPD matrices, very few solutions explicitly mine the local geometrical information in deep SPD feature representations. 
Given the great success of local mechanisms in Euclidean methods, we argue that it is of utmost importance to ensure the preservation of local geometric information in the SPD networks.
We first analyse the convolution operator commonly used for capturing local information in Euclidean deep networks from the perspective of a higher level of abstraction afforded by category theory. 
Based on this analysis, we define the local information in the SPD manifold and design a multi-scale submanifold block for mining local geometry.
Experiments involving multiple visual tasks validate the effectiveness of our approach.
The supplement and source code can be found in https://github.com/GitZH-Chen/MSNet.git.
\end{abstract}

\section{Introduction} \label{sec:intro}
Symmetric Positive Definite (SPD) matrices have shown great success in many scientific areas, like medical image analysis\cite{chakraborty2020manifoldnet, chakraborty2018statistical}, elasticity \cite{guilleminot2012generalized}, signal processing \cite{hua2017matrix, brooks2019riemannian}, machine learning \cite{kulis2006learning,harandi2012sparse}, and computer vision \cite{chakraborty2020manifoldnorm,zhang2020deep,zhen2019dilated,huang2017riemannian,nguyen2021geomnet,harandi2018dimensionality}.
However, the non-Euclidean nature of SPD matrices precludes the use of a wide range of data analysis tools, the applicability of which is confined to Euclidean metric spaces. 
Motivated by this problem, a number of Riemannian metrics have been introduced, such as Log-Euclidean metric (LEM) \cite{arsigny2005fast}, Affine-Invariant Riemannian metric (AIM) \cite{pennec2006riemannian}, Log-Cholesky metric (LCM) \cite{lin2019riemannian}. 

With these well-studied Riemannian metrics, some Euclidean techniques can be generalized into SPD manifolds. 
Some approaches adopted approximation methods that locally flatten the manifold by identifying it with its tangent
space\cite{huang2015log}, or by projecting the manifold into a Reproducing Kernel Hilbert Spaces (RKHS)\cite{chen2021hybrid, harandi2012sparse, wang2012covariance}.
However, these methods tend to distort the geometrical structure communicated by the data. 
To address this issue, \cite{harandi2018dimensionality, gao2019robust} proposed direct learning algorithms on SPD manifolds.
In addition, inspired by the significant progress achieved by deep learning \cite{lecun1998gradient, krizhevsky2012imagenet}, \cite{huang2017riemannian, zhang2020deep, brooks2019riemannian,chakraborty2018statistical,chakraborty2020manifoldnet,chakraborty2020manifoldnorm,nguyen2021geomnet,zhen2019dilated,wang2022spddeep,wang2022dreamnet} attempted to build deep Riemannian networks for non-linear representation learning on SPD matrices. 

Nevertheless, many existing deep SPD networks treat the SPD features as a global representation. 
Given the success of multi-scale features in both conventional feature design \cite{belongie2002shape, lowe2004distinctive} and deep learning \cite{szegedy2015going, krizhevsky2012imagenet}, it should be rewarding to investigate local mechanism in Riemannian neural networks. 
Accordingly, in this paper, we develop a deep multi-scale submanifold network designed to capture the informative local geometry in deep SPD networks. 
To the best of our knowledge, this is the first work to successfully mine the local geometric information on SPD manifolds.

As convolution is one of the most successful techniques for dealing with local information in traditional deep learning, we first analyze its mathematical essence from the perspective of category theory, to identify the universal property which is transferable to manifolds.
We proceed to define the local information in the category of SPD manifolds and propose multi-scale submanifold blocks to capture both holistic and local geometric information.
In summary, our contributions are three-fold:
1).
a theoretical guideline is developed for the Riemannian local mechanism.
2).
local patterns in Riemannian manifolds are rigorously defined.
3).
a novel multi-scale submanifold block is proposed to capture vibrant local statistical information on the SPD networks. 

\section{Related Work} \label{sec:re_work}
To take advantage of deep learning techniques, some effort has been made to generalize Euclidean deep learning into a Riemannian one. 
\cite{huang2017riemannian} designed a densely connected feedforward network on SPD matrices with a procedure involving a slice of spectral layers.
\cite{chakraborty2020manifoldnet} proposed a theoretical framework to fulfil convolution network on Riemannian manifolds, where each 'pixel' of the input tensor is required to be a manifold-valued point.
However, different from Euclidean convolution, none of these methods pay attention to the local information in a single SPD matrix. 
In contrast, \cite{zhang2020deep} proposed an SPD transformation network for action recognition.
They designed an SPD convolutional layer, which is similar to the Euclidean convolution except that the convolutional kernels are required to be SPD.
Note that the square matrices covered by a sliding window might not be SPD. 
Therefore, local geometry might be distorted or omitted, undermining the efficacy of their proposed network. 
In contrast, in our approach, the proposed mechanisms can faithfully preserve local information.
A multi-scale representation is further adopted, which captures different levels of statistical information.
We expect that Riemannian networks can benefit from this comprehensive statistical information.

\section{Preliminaries} \label{sec:pre}
To develop our proposed method, category theory and regular submanifolds are briefly reviewed.
Due to the page limit, others such as differential manifolds, the geometry of SPD manifolds, and our backbone network, SPDNet, are presented in the supplement. 

\subsection{Foundations of Category Theory}
Category theory, which is similar to object-oriented programming in computer science, studies the universal properties and mathematical abstractions shared by different domains.
\begin{definition} \label{def:category}
A category $\calC$ consists of a collection of elements, called objects, denoted by $\obj(\calC)$, and a set $\mor(A, B)$ of elements, called morphisms from $A$ to $B$, for any two objects $A, B \in \obj(\calC)$.
Morphisms should satisfy the below three axioms:
\begin{itemize}
    \item composition: 
    given any $f \in \mor(A,B)$ and $ g \in \mor(B,C)$, the composition $h = g \circ f \in \mor(A,C)$ is well-defined.
    \item
    identity: 
    for each object $A$, there is an identity morphism $\idmap_{A} \in \mor(A, A)$ such that for any $f \in \mor(A, B)$ and $g \in \mor(B, A)$,
    \begin{equation}
        f \circ \idmap_{A}=f, \idmap_{A} \circ g=g ;
    \end{equation}
    \item
    associative: 
    for $f \in \mor(A, B), g \in \mor(B, C)$, and $h \in \mor(C, D)$,
    \begin{equation}
    h \circ(g \circ f)=(h \circ g) \circ f .
    \end{equation}
\end{itemize}
The set of all the morphisms in $\calC$ is denoted as $\mor(\calC)$.
\end{definition}

Let us take the linear space, which is widely studied in pattern recognition, as an example.
In this category, the objects are linear spaces and morphisms are linear homomorphisms.
More details are introduced in \cite[$\S$ 10]{loring2011introduction}.

Category theory enables us to develop a mathematical abstraction of operations in one category and generalize them into another category.
In this paper, we will rely on this theory to derive local mechanisms in Riemannian manifolds from Euclidean counterparts.

\subsection{Regular Submanifolds}
Regular submanifolds \cite[$\S$ 9]{loring2011introduction} of manifolds generalize the idea of subspace in the Euclidean space.
In the smooth category, since manifolds are locally Euclidean, submanifolds are defined locally.
\begin{definition} \label{def:submanifold}
A subset $\calS$ of a smooth manifold $\calN$ of dimension $n$ is a regular submanifold of dimension $k$ if for every $p \in \calS$ there is a coordinate neighbourhood $(U, \phi)=\left(U, x^{1}, \ldots, x^{n}\right)$ of $p$ in the maximal atlas of $\calN$ such that $U \cap \calS$ is defined by the vanishing of $n-k$ of the coordinate functions.
\end{definition}

\section{Proposed Method}
\label{method}
\begin{figure*}
  \centering
  \includegraphics[width=0.8\linewidth]{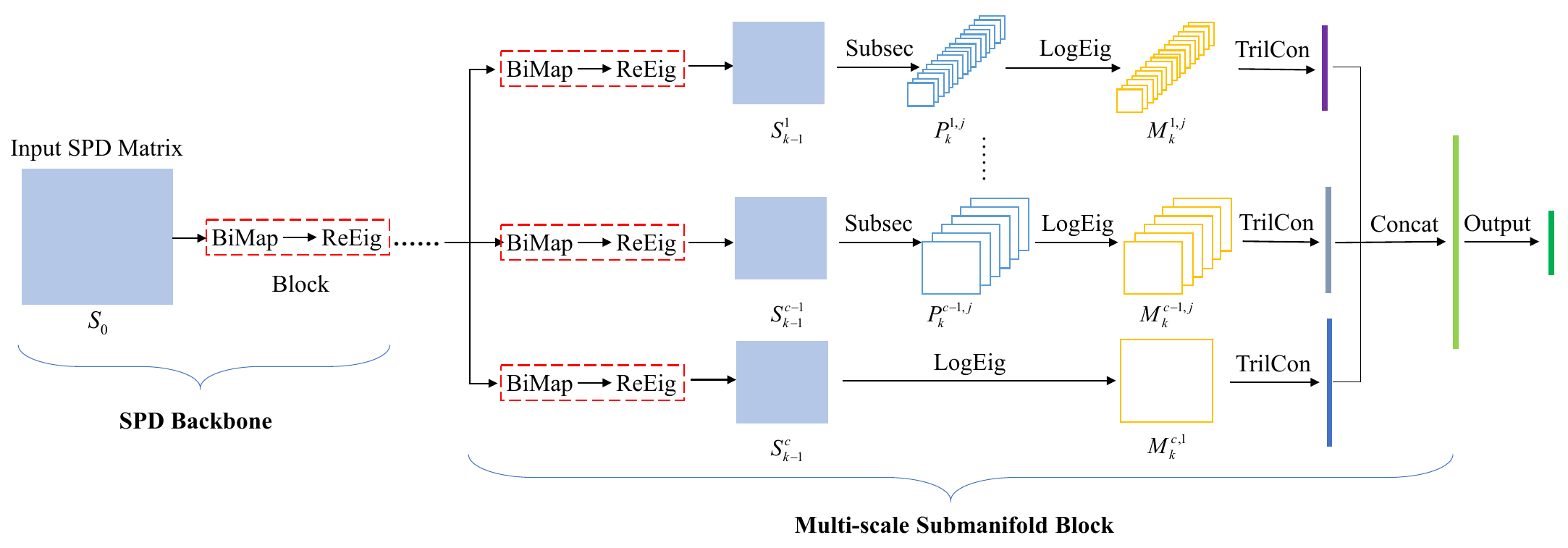}
  \caption{ Illustration of the proposed Multi-scale Submanifold Network (MSNet). 
  We first employ SPDNet \cite{huang2017riemannian} as our backbone to extract lower dimensional, yet more discriminative SPD feature representations.
  Then in each branch, a BiMap-ReEig block is exploited to obtain SPD representations $S_{k-1}^i$, where $k-1$ and $i$ are layer and channel index respectively.
  We propose a submatrix selection, denoted as SubSec, to result in $P_k^{i,1} \cdots P_k^{i,n^i}$ along the $i^{th}$ channel for local manifold feature learning, where $n^i$ is the number of selected submanifolds in the $i^{th}$ channel.
  Next, LogEig layer is applied to map each  submatrix feature into a Euclidean space, \textit{i.e.,} $M_k^{i,j} = \log(P_k^{i,j})$.
  Then, we apply TrilCan to go through a process of extracting a lower triangular matrix, vectorization, and concatenation.
  Finally, we concatenate all the vectors from the different branches with a Concat layer, followed by any regular output layers like the softmax layer.
  Note that the SPD matrix itself can also be viewed as a trivial submanifold, encoding global information, and hence the bottom branch is exploited to capture global information, which is how SPDNet works.}
  \label{fig:framework}
\end{figure*}
In this section, we will introduce our method in detail.
As Euclidean convolution is one of the most successful local mechanisms, we first analyze this operation from the perspective of category theory to uncover the mathematical essence of the Euclidean local mechanism. 
Under this analysis, we proceed to define the local pattern in the SPD manifold.
Finally, we introduce a multi-scale local mechanism to capture fine-grained local information.
The proposed network is conceptually illustrated in \figref{fig:framework}.

\subsection{Analysis of Euclidean Convolution} \label{sec:analysis_conv}
Since the convolution is an operation in the category of linear space, we first consider it from the perspective of linear algebra.
Then we will proceed with the analysis via category theory to derive the general properties that can be transferred into the SPD manifold.

To avoid tedious subscripts, we consider an example whose input and output are single channel tensors. 
In this case, only one kernel filter is involved, but the following analysis can also be readily generalized into an arbitrary number of channels.
In convolution networks, the tensor feature and the $i^{th}$ receptive field can be viewed as the linear space $V$ and subspace $V_i$, respectively.
The process that the $i^{th}$ receptive field is reduced into a real scalar by a specific kernel filter can be deemed as a linear mapping $f_i: V_i \rightarrow M_i$.
Note that $M_i$ is a trivial one-dimensional linear space, $\mathbb{R}$. 
After convolution, each receptive field is reduced into a real number and these scalar elements are concatenated to build a new tensor feature.
This process can be more generally described by the notion of direct sum ``$\oplus$'' \cite{roman2005advanced} in linear algebra, \textit{i.e.,} $M = M_1 \oplus \cdots \oplus M_n$.
Not that the direct sum, ``$\oplus$'' can be intuitively understood as a counterpart of the union in the set theory.
(see supplement for more details)
The above analysis leads to the following abstraction of the convolution operation. 
\begin{props} \label{prop:conv_linear}
For a given linear space $V$ of dimension $d \times d$, $n$ linear subspaces $V_1, V_2,\cdots, V_n$ of dimension $k \times k$ are selected and a linear function $f_i(\cdot): V_i \rightarrow M_i$ is performed in each of them to extract local linear information. 
The resulting linear spaces $M_1,\cdots,M_n$ are combined into a final linear space $M$ by direct sum, \textit{i.e.},$M = M_1 \oplus \cdots \oplus M_n$.
\end{props}

To discover a more general property of Euclidean convolution, we further analyze it by category theory.
To this end, we can simply substitute the linear algebra terms with category language following the axioms of category theory.
In detail, linear space $V$, subspace $V_i$ and linear function $f_i$ can be more generally described as object $A$, sub-object $A_i$ and morphism $f_i$.
Besides, we notice that each subspace $V_i$ shares the same dimensionality, which indicates $V_1, \cdots, V_n$ are equivalent in the sense of linear space.
This suggests that the extracted sub-objects $A_1, \cdots, A_n$ should be isomorphic.
However, not all categories share the idea of the direct sum. For example, the categories, known as group, ring and field, do not have this kind of operation. 
Therefore, the combination strategies vary in different categories.
Now, we could obtain a more general description of convolution by category theory.
\begin{props} \label{prop:conv_cat}
In a category $\mathcal{C}$, for an object $A \in Obj(\mathcal{C})$, we extract $n$ isomorphic sub-objects from $A$, denoted by $A_{1}, A_{2}, \ldots, A_{n}$. 
Then morphism $f_{i} \in \operatorname{Hom}\left(A_{i}, B_{i}\right)$ is applied to each sub-object to map it into a resulting object $B_{i}$.
The resulting objects $B_1,\cdots,B_{i}$ are combined into a final object $B \in Obj(\mathcal{C})$ according to certain principles.
\end{props}

With the \propsref{prop:conv_cat} as an intermediary, we can generalize the convolution into manifolds theoretically. 
Specifically, the object, sub-object and morphism in manifolds are manifolds, submanifolds and smooth maps respectively.
\begin{props} \label{prop:conv_manifold}
For a manifold $\mathcal{M}$, we extract $n$ isomorphic submanifolds $\mathcal{M}_1, \mathcal{M}_2, \ldots,\mathcal{M}_n$ and map each one of them by a smooth maps $f_i(\cdot): \mathcal{M}_i \rightarrow \mathcal{M}_i'$.
The resulting manifolds $\mathcal{M}_i'$ are aggregated into a final manifold $\mathcal{M}'$ according to certain principles.
\end{props}

As a summary of the above discussion, we obtain the following important insights about the Riemannian local mechanism.
First of all, the local patterns in manifolds are submanifolds.
Secondly, not all submanifolds are involved, and all the selected submanifolds could be isomorphic.
Lastly, a specific way of aggregating submanifolds should be elaborately designed according to the axioms of manifolds.

\subsection{Submanifolds in SPD Manifolds}

\propsref{prop:conv_manifold} demonstrates that local patterns in manifolds are submanifolds.
In this subsection, we will identify the submanifolds in the specific SPD manifolds.
Briefly speaking, in the category of SPD manifolds, submanifolds should further be SPD manifolds.
This constraint can be fulfilled by principle submatrices.

For a clearer description, let us make some notifications first.
Denote k-fold row and column indices as $\calI=\{i_1,\cdots,i_k\}$ and $\calJ=\{j_1,\cdots,j_k\}$.
For a set of real square matrices $\bbR{n \times n}$, we denote $(\bbR{n \times n})_{\calI,\calJ}$ as the set of submatrices, which are obtained by remaining rows $\calI$ and columns $\calJ$.
If $\calI=\calJ$, then $(\bbR{n \times n})_{\calI,\calI}$ is the set of principal submatrices, abbreviated as $(\bbR{n \times n})_{\calI}$.

As we discussed before, subspaces are sub-objects in the category of linear algebra.
For a set of real square matrices $\bbR{n \times n}$, any set of $k \times k$ submatrices, $(\bbR{n \times n})_{\calI,\calJ}$, forms a subspace of $\bbR{n \times n}$.
However, things would be different for SPD manifolds.
Linear algebra tells us that an arbitrary submatrix of an SPD matrix might not be SPD, and even not be symmetric, while principle submatrices are always SPD.
Although we can readily prove that $(\spd{n})_{\calI,\calJ}$ can be viewed as a regular submanifold of the SPD manifold $\spd{n}$, $(\spd{n})_{\calI,\calJ}$ might not be a SPD manifold.
This could cause some inconsistency, since in the specific category of SPD manifolds, objects should always be SPD manifolds.
In addition, $(\spd{n})_{\calI}$ can be viewed as a regular submanifold of $\spd{n}$.
The above discussion is formalized by the following theorem.
(Proof is presented in the supplement.)
\begin{theorem} \label{thm:submanifolds}
For an SPD manifold $\spd{n}$, the set of principal submatrices $(\spd{n})_{\calI}$ is an SPD manifold and can be embedded into the $\spd{n}$ as a regular submanifold.
In addition, for any proper indices $\calI$ and $\calJ$ satisfying $|\calI| = |\calJ|$, $(\spd{n})_{\calI}$ is isomorphic to $(\spd{n})_{\calJ}$.
\end{theorem}
Now, we have identified that the local pattern in the specific SPD manifolds are principal submatrices.

\subsection{Multi-scale Submanifold Block} \label{sec:ms_block}
Because of the analysis in \secref{sec:analysis_conv}, there are two factors we should consider when designing our submanifold block, \textit{i.e.}, the rule for selecting isomorphic submanifolds and the way of aggregating them.
In the following, we first discuss the details of selecting submanifolds in the SPD manifold.
Then we proceed to introduce our multi-scale submanifold block, which fuses submanifolds via a divide-aggregation strategy.
\subsubsection{The Principles for Selecting Submanifolds} \label{sec:pre_sel}
\begin{figure}[t]
  \centering
   \includegraphics[width=0.8\linewidth]{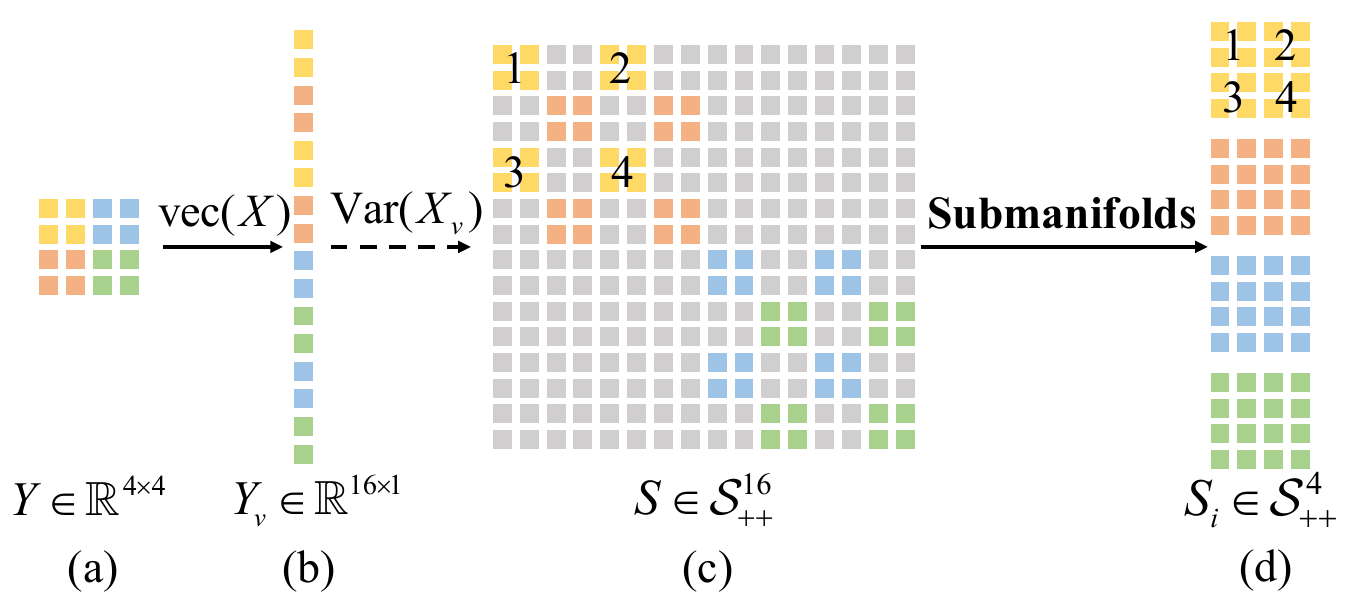}
   \caption{
    Illustration of the process of selecting principal submatrices. 
   (a) We deem an $S\in \mathcal{S}_{++}^{16}$ as a covariance from an imaginary $4 \times 4$ random matrix $Y$.
   We use a $2 \times 2$ sliding window with skip of $2$ on $X$ to obtain the corresponding position index.
   We use four kinds of colour to denote the four regions of interest.
   (b) shows the corresponding indexes of the four regions in vectorized $Y_v$.
   (c) Then we can find the corresponding region covariance matrices for the four regions from $S$.
   (d) The region covariance matrices $S_i \in \mathcal{S}_{++}^4$ corresponding to these local regions are the submanifolds we select. 
   }\label{fig:pre_sel}
\end{figure} 
\thmref{thm:submanifolds} reveals that the isomorphic submanifolds in SPD manifolds are the sets of principal submatrices of the same size.
Note that incorporating all the sub-objects might not be an optimal solution, since it introduces redundant information with cumbersome computation. 
In conventional convolution, sub-objects are selected following the concept of the receptive field. 
In terms of specific SPD manifolds, the number of principal submatrices grows combinatorially, which could cause a dimensionality explosion.
For instance, if we select all the principal submatrices of $4 \times 4$ from a $16 \times 16$ SPD matrix, which seems like a trivial example,
the total number of selected submatrices would be $C_{16}^4 = 1820$.
In most cases, this huge number of small SPD submatrices would not be manageable.
Therefore, it is crucial to select principal submatrices to construct submanifolds.
To this end, we introduce a method for selecting principal submatrices.

In probability theory, the distribution information of a $d \times d$ random matrix $X$ can be conveyed by its covariance matrix $S$ of $d^2 \times d^2$, where $S = \operatorname{Var}( \operatorname{vec}(X))$ and $\operatorname{vec}(\cdot)$ denotes vectorization.
The covariance of a $k \times k$ local region in $X$ corresponds to a $k^2 \times k^2$ principal submatrix of $S$.
If we focus on multiple $k \times k$ local regions, we then can extract a series of principal submatrices of $k^2 \times k^2$.
Considering that an SPD matrix is often defined by covariance in computer vision, we can follow this probabilistic clue to select principal submatrices.

Specifically, according to statistics, for a given image set consisting of images $X_1, \cdots, X_m $ of the size of $n \times n$, it can be viewed as $m$ samples from a population, an $n \times n$ random matrix $X$.
Then we estimate population covariance by sample covariance and model it into an $n^2 \times n^2$ SPD matrix $S_0$.
After forward passing the deep SPD networks, we obtain a lower-dimensional discriminative SPD matrix $S$ of $d^2 \times d^2$ via the network mapping denoted by $\phi(\cdot): \mathcal{S}_{++}^{n^2} \rightarrow \mathcal{S}_{++}^{d^2}$.
We hypothesize that there is an implicit mapping $\psi(\cdot): \mathbb{R}^{n \times n} \rightarrow \mathbb{R}^{d \times d}$ to transform the random matrix $X$ into another one $\psi(X)$ and $S$ is the covariance of $\psi(X)$.
We denote $\psi(X)$ as $Y$ for simplicity and then the size of $Y$ is $d \times d$.
If we focus on the local region covariance of $Y$, then we can extract a series of principal submatrices from $S$.
Besides, the number of submatrices extracted in this way is much smaller than the combinatorial number.

More specifically, we view a $d^2 \times d^2$ SPD matrix $S$ as a covariance from a $d \times d$ random matrix $Y$, 
The covariance matrix associated with a $k \times k$ receptive field of the matrix $Y$ corresponds to a $k^2 \times k^2$ principal submatrix of $S$.
When moving a $k \times k$ sliding window by a step equal to $s$ in $Y$ (to obtain the position index), we can select $(\frac{d-k}{s}+1)^2$ principal submatrices of $S$ accordingly. 
Obviously, the submatrices we select are still SPD and the number of them is much smaller than $C_{d^2}^{k^2}$.
Besides, it will encode the geometrical information conveyed by the local system in the category of SPD manifold.
\figref{fig:pre_sel} provides a conceptual illustration of the process of selecting submanifolds.

\subsubsection{Multi-scale Mechanism} \label{sec:Muti_mechanism}
In fact, in a deep SPD network, hidden SPD feature contains statistically compact information. 
Each element of the hidden feature reflects the region correlation in the original video.
By extracting principal submatrices, we focus on the correlation among multiple local regions.
In this way, we can capture the local semantic information in a statistical form.
We anticipate that the classification can benefit from this compact local statistical information. 

Furthermore, inspired by Res2Net \cite{gao2019res2net}, which attempts to capture multi-scale features in one single residual block, we design a multi-channel mechanism to obtain local manifold information at diverse granularity levels. 
In detail, for an SPD feature, various BiMap-ReEig layers are applied to obtain a low dimensional SPD matrix, by:
\begin{align}
    &\text{BiMap: } S_{k-1}^i =W_{k-1}^i S_{k-2} W_{k-1}^{i T},\\
    &\text{ReEig: } S_k^i = U_{k-1}^i\max(\epsilon I,\Sigma_{k-1}^i)U_{k-1}^{i T},     \label{eq:reeig}
\end{align}
where all $W_{k-1}^i$ are of the same size, and \eqref{eq:reeig} is an eigenvalues function.

It is expected that in each channel, the primary learnt local geometry $S_k^i$ varies.
To capture it, submatrices with specific $k_i \times k_i $ dimensions are extracted from $S_k^i$.
In this way, rich statistical local information at different granularity levels is extracted by our network.
Besides, as an SPD matrix itself can be viewed as a trivial submanifold, we also incorporate a channel capturing the global information.
In this way, not only holistic geometry, but also local geometry are jointly captured in the category of SPD manifold. 

\subsubsection{Submanifolds Fusion Strategies}

If we follow the \propsref{prop:conv_manifold} rigorously, it would result in a major computational burden, in view of the complex structure of the Riemannian manifold.
Therefore, we fuse this multi-scale information in an approximate way instead.
Since the SPD manifold $\mathcal{S}_{++}^d$ is diffeomorphic to the Euclidean space $\mathcal{S}^d$ of symmetric matrices by matrix logarithm, we can fuse the multi-scale information in a comparatively simple space, \textit{i.e.}, $\mathcal{S}^d$, which is a well-known technique for tackling the problems of data analysis on SPD manifolds \cite{huang2015log}.

To be more specific, we first map each principal submatrix into a real symmetric matrix by matrix logarithm, $M_k^{i j} = \log(P_{k}^{i j})$, where $k,i$ and $j$ represent $k^{th}$ layer,$i^{th}$ channel, and $j^{th}$ submatrix respectively.
Considering that the dimensionality of the Euclidean space formed by symmetric matrices $\mathcal{S}^d$ is $\frac{d(d+1)}{2}$ , for each $M_k^{i j}$, we exploit its lower triangular part to further mitigate the computational burden.
Since at this point the features lie in a Euclidean space, concatenation can be applied to fuse them.
This process can be formally described as:
\begin{equation} \small
    V_{k}^{i}=\operatorname{concat}(\operatorname{vec}(\operatorname{tril} (M_{k}^{i 1})), \cdots, \operatorname{vec}(\operatorname{tril}(M_{k}^{i n^{i}})))
\end{equation}
where $\operatorname{vec}(\cdot)$ means vectorization.

After we fuse the local information in each channel, we concatenate the feature vectors to aggregate the multi-scale information in different channels.

\section{Experiments}
We evaluate the proposed MSNet in three challenging visual classification tasks: video-based action recognition with the Cambridge-Gesture (CG) \cite{kim2008canonical} and the UCF-101 \cite{soomro2012ucf101} datasets, and skeleton-based action recognition with the First-Person Hand Action (FPHA) \cite{garcia2018first} dataset, respectively.
We simply employ a fully connected layer, softmax layer and cross-entropy as our output layer as \cite{huang2017riemannian, huang2018building}. 
For training our MSNet, we use an i5-9400 (2.90GHz) CPU with 8GB RAM.
\subsection{Implementation Details}
The SOTA Riemannian learning competitors include: 
\textit{1). General methods for SPD learning:}
Covariance Discriminative Learning (CDL) \cite{wang2012covariance},
SPD Manifold Learning (SPDML-AIM, SPDML-Stein) \cite{harandi2018dimensionality} and 
 Log-Euclidean Metric Learning (LEML) \cite{huang2015log};
\textit{2). General methods for Grassmann learning: }    
Grassmann Discriminant Analysis (GDA) \cite{hamm2008grassmann} and 
Projection Metric Learning (PML) \cite{huang2015projection};
\textit{3). Hybrid Riemannian manifold learning methods: }      
Hybrid Euclidean-and-Riemannian Metric Learning (HERML) \cite{huang2015face} and 
Multiple Manifolds Metric Learning (MMML) \cite{wang2018multiple};
\textit{4). Riemannian deep methods: }        
    SPD Network (SPDNet) \cite{huang2017riemannian},
    SymNet \cite{wang2021symnet},
    Grassmannian Network (GrNet) \cite{huang2018building}.
All the comparative methods are carefully re-implemented by the source codes and fine-tuned according to the original papers.

To further evaluate the effectiveness of our algorithm, we also compare our MSNet with conventional SOTA hand pose estimation methods on the FPHA dataset.
These approaches include 
Lie Group \cite{vemulapalli2014human}, 
Hierarchical Recurrent Neural Network (HBRNN) \cite{du2015hierarchical},
Jointly Learning Heterogeneous Features (JOULE) \cite{hu2015jointly},  
Convolutional Two-Stream Network (Two stream) \cite{feichtenhofer2016convolutional},
Novel View \cite{rahmani20163d}, 
Transition Forests (TF) \cite{garcia2017transition}, 
Temporal Convolutional Network (TCN) \cite{kim2017interpretable}, 
LSTM \cite{garcia2018first}
and Unified Hand and Object Model \cite{tekin2019h+}. 
Besides, we also compare our approach against Euclidean network searching methods, DARTS\cite{liu2018darts} and FairDARTS \cite{chu2020fair}, following the setting in \cite{sukthanker2021neural} by viewing SPD logarithm maps as Euclidean data.

We study five configurations, \textit{i.e.}, MSNet-H, MSNet-PS, MSNet-AS, MSNet-S and MSNet-MS to further evaluate the utility of our proposed network.
In detail, MSNet-MS with two BiMap-ReEig layers is ${S}_{0} \rightarrow f_{b}^{(1)} \rightarrow f_{r}^{(2)} \rightarrow f_{b}^{(3)} \rightarrow f_{r}^{(4)} \rightarrow f_{m}^{(5)}  \rightarrow f_{f}^{(6)} \rightarrow f_{s}^{(7)}$, where $f_b^{(k)}, f_r^{(k)}, f_{m}^{(k)}, f_f^{(k)},f_s^{(k)}$ represent $k^{th}$ layers of BiMap, ReEig, multi-scale submanifold, FC, and softmax-cross-entropy, respectively.
Note that apart from the two BiMap-ReEig blocks in the backbone, in each branch of $f_{m}^{(k)}$, there is a BiMap-ReEig block as well, as illustrated in \figref{fig:framework}.
Besides, as the whole SPD matrix can be viewed as a trivial submanifold, we use MSNet-H to denote that we only extract holistic information. 
Though it is similar to SPDNet, only the lower triangular part is exploited for classification in our framework, alleviating the computational burden.
To study the utility of proper submanifolds, we use MSNet-PS to represent that we extract all kinds of proper submanifolds according to our principles except the trivial ones.
To see whether over-loaded submanifolds would bring about redundant information, we build MSNet-AS to extract all the submanifolds including the trivial ones.
MSNet-S denotes that we only utilize the proper submanifolds in the corresponding MSNet-MS except for the trivial one.

In the experiments, we simply set ``step'' size equal to $1$.
The above models of our MSNet and SPDNet share the same learning mechanism as follows.
The initial learning rate is $\lambda = 1e^{-2}$ and reduced by $0.8$ every $50$ epochs to a minimum of $1e^{-3}$.
Besides, the batch size is set to 30, and the weights in BiMap layers are initialized as random semi-orthogonal matrices.
For activation threshold in ReEig and dimension of transformation matrices in BiMap, we first search the optimal settings by our backbone SPDNet and then employ the same settings to our MSNet.

\subsection{Datasets and Settings}
\begin{table}
  \small
  \centering
    \begin{tabular}{cccc}
    \toprule
    Dataset & CG    & FPHA  & UCF-sub \\
    \midrule
    BiMap Settings & {100,80,50,25} & {63,56,46,36} & {100,80,49} \\
    Submanifolds & $ 2^2,3^2,4^2,5^2$ & $5^2,6^2$ & $2^2,6^2,7^2$ \\
    Learning Rates & $1e^{-5}$ & $1e^{-4}$ & $1e^{-5}$ \\
    Epochs & 500   & 3500  & 500 \\
    \bottomrule
    \end{tabular}
  \caption{Configurations of MSNet on three datasets. Note that 100,80,50,25 means $100 \times 80, 80 \times 50, 50 \times 25$.}
  \label{tab:parameters}
\end{table} 

To evaluate our method when facing limited data, experiments are carried out on the CG \cite{kim2008canonical} dataset.
It consists of $900$ video sequences covering nine kinds of hand gestures.
For this dataset, following the criteria in \cite{chen2020covariance}, we randomly select $20$ and $80$ clips for training and testing per class, respectively.
For evaluation, we resize each frame into $20 \times 20$ and obtain the grey scale feature.
To further facilitate our experiment, we reduce each frame dimension to $100$ by PCA. 
Then we compute the covariance matrix of size $100 \times 100$ to represent each video.

We employ the popular FPHA \cite{garcia2018first} dataset for skeleton-based action recognition. 
It includes 1,175 action videos of 45 different action categories. 
For a fair comparison, we follow the protocols in \cite{wang2021symnet}.
In detail, we use 600 action clips for training and 575 for testing and each frame is vectorized into a 63-dimensional feature vector with the provided 3D coordinates of 21 hand joints.
Then we obtain a $63 \times 63$ covariance representation for each sequence.

To assess the utility of our method in the task of relatively large scale, UCF-101 \cite{soomro2012ucf101} dataset is exploited, which is sourced from YouTube, containing 13k realistic user-uploaded video clips of 101 types of action. 
To facilitate our experiment, 50 kinds of action are randomly selected, each of which consists of 100 clips. 
We call this dataset UCF-sub in the following. 
As we did in the CG dataset, we exploit the grey scale feature and reduce the dimension of each frame to 100 by PCA.
The seventy-thirty-ratio (STR) protocol is exploited to build the gallery and probes.

The configurations on three datasets are listed in \tabref{tab:parameters}.
\subsection{Analysis}
\begin{table}[!t]
\small
\centering
\begin{tabular}{cccccc}
\toprule
Methods & Year & Colour & Depth & Pose & Acc. \\
\midrule
Lie Group  & 2014 & \ding{55} & \ding{55} & \ding{51} & 82.69\\
HBRNN  & 2015 & \ding{55} & \ding{55} & \ding{51} & 77.40 \\
JOULE  & 2015 & \ding{51} & \ding{51} & \ding{51} & 78.78 \\
Two stream  &2016 & \ding{51} & \ding{55} & \ding{55} & 75.30 \\
Novel View  & 2016 & \ding{55} & \ding{51} & \ding{55} & 69.21 \\
TF  & 2017 & \ding{55} & \ding{55} & \ding{51} & 80.69\\
TCN  & 2017 & \ding{55} & \ding{55} & \ding{51} & 78.57 \\
LSTM  & 2018 & \ding{55} & \ding{55} & \ding{51} & 80.14\\   
H+O  & 2019 & \ding{51} & \ding{55} & \ding{55} & 82.43\\
DARTS  & 2018 & \ding{55} & \ding{55} & \ding{51} & 74.26\\
FairDARTS  & 2020 & \ding{55} & \ding{55} & \ding{51} & 76.87\\
SPDML-AIM  & 2018 & \ding{55} & \ding{55} & \ding{51} & 76.52\\
HERML  & 2015 & \ding{55} & \ding{55} & \ding{51} & 76.17\\
MMML  & 2018 & \ding{55} & \ding{55} & \ding{51} & 75.05\\
SPDNet  & 2017 & \ding{55} & \ding{55} & \ding{51} &85.57\\
GrNet  & 2018 & \ding{55} & \ding{55} & \ding{51} & 77.57\\
SymNet  & 2021 & \ding{55} & \ding{55} & \ding{51} & 82.96\\
\midrule
MSNet-H &   & \ding{55} & \ding{55} & \ding{51} & 85.74\\
MSNet-PS &   & \ding{55} & \ding{55} & \ding{51} & 80.52\\
MSNet-AS &   & \ding{55} & \ding{55} & \ding{51} & 82.26\\
MSNet-S &   & \ding{55} & \ding{55} & \ding{51} & 86.61\\
\bfseries MSNet-MS
  & & \ding{55} & \ding{55} & \ding{51} & \textbf{87.13} \\
\bottomrule
\end{tabular}
\caption{Recognition Results (\%) on the FPHA Dataset.}
\label{tab:fpha}
\end{table}

\begin{table}
  \small
  \centering
  \begin{tabular}{*{3}{c}}
    \toprule
     Method &  CG  &  UCF-sub  \\
    \midrule
     GDA  & 88.68 & 43.67\\
     CDL  & 90.56 & 41.53 \\
     PML  & 84.32 & 50.60 \\
     LEML  & 71.15 & 44.67 \\
     SPDML-Stein  & 82.62 & 51.40\\
     SPDML-AIM  & 88.61 & 51.13 \\
     HERML  & 88.94 & NA \\
     MMML & 89.92 & NA \\
     GrNet & 85.69 & 35.80 \\
     SPDNet & 89.03 & 59.93 \\
     SymNet & 89.81 & 56.73 \\
     \midrule
     MSNet-H & 89.03 &  58.27\\
     MSNet-PS & 90.14 & 57.73 \\
     MSNet-AS & NA & 58.33 \\
     MSNet-S & 90.14 & 59.40 \\
     MSNet-MS & \textbf{91.25} & \textbf{60.87}\\
    \bottomrule
  \end{tabular}
  \caption{Performance (\%) on the CG and UCF-sub datasets.}
  \label{tab:results}
\end{table}

\begin{figure}[t]
  \centering
  \includegraphics[width=0.8\linewidth]{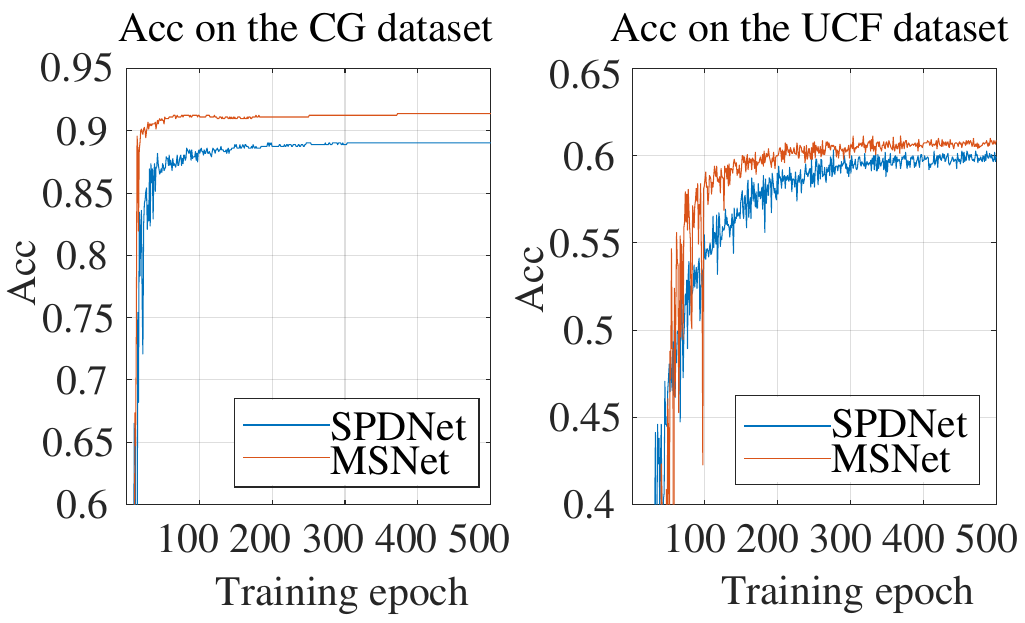}
  \caption{Accuracy curve of the proposed MSNet against SPDNet on the CG and UCF-sub datasets.}\label{fig:convergence}
\end{figure}

As reported in \tabref{tab:fpha} and \tabref{tab:results}, our proposed MSNet-MS outperforms all the other competitors on three datasets. 
Note that in a relatively large dataset, some shallow learning methods, such as MMML and HERML, are barely feasible in view of their time-consuming optimization.
\figref{fig:convergence} shows the convergence behaviour on the CG and UCF-sub datasets.
There are several interesting observations worth discussing.

Firstly, although hybrid Riemannian methods, like HERML and MMML,  can take advantage of complementary information encoded in different manifold and thus surpass some deep methods occasionally, our method consistently outperforms them.
This verifies that information encoded in submanifolds is of great importance and could be beneficial for classification.

Secondly, on the CG and FPHA datasets, MSNet-H achieves almost the same performance as SPDNet, while on the UCF-sub dataset,  MSNet-H is inferior to SPDNet.
Although as we discussed in \secref{sec:pre_sel}, the lower triangular part of a symmetric matrix is mathematically equivalent to itself, the non-convexity of optimization on deep learning might cause some empirical deviation.
However, extracting lower triangular, which is the way we exploit, could alleviate the computational burden and thus enhance the scalability of the output layer.
For instance, if a deep network is used as an output classifier, it would be more efficient to halve the dimensionality of the input vector.
What's more, our method, \textit{i.e.}, MSNet-MS, surpasses the backbone network, SPDNet, in all three datasets.
This suggests that the underlying statistical information in submanifolds could contribute to the visual perception, leading to a better classifier.
Therefore, efforts should be paid to mine the information in submanifolds.
However, over-saturated efforts in submanifolds might undermine discriminability.
This idea is justified by the generally inferior performance of MSNet-AS, which selects all the submanifolds, to MSNet-MS. 

Thirdly, although proper efforts should be made for submanifolds, they might vary for different tasks. 
On the CG dataset, the best performance is achieved when we use all kinds of submanifolds.
This might be attributed to the particularity of the dataset.
In detail, on this dataset, the background and the foreground are relatively monotonous, and the difference between them is apparent.
Therefore, statistical information at diverse granular levels encoded in different submanifolds could contribute to the classification.
However, on the FPHA and UCF-sub datasets, the large variance of appearance makes us cautious to select submanifolds to waive the statistically dispensable information. 

More importantly, although local statistical information is of great value for visual classification, never can we neglect the importance of holistic information. 
Specifically, as we can see, MSNet-MS is superior to MSNet-S. 
The sole difference between these two configurations is that MSNet-MS uses the global information, while MSNet-S does not.
The consistent phenomenon can be observed in the comparison between MSNet-AS and MSNet-PS.
This indicates that only when we combine the optimal local information together with global information, could we make the best of statistical information.

It takes about 1.29s, 2.67s and 34.50s per epoch to train our MSNet-MS on the CG, FPHA and UCF-sub datasets, respectively, while training SPDNet takes 0.53s, 1.53s and 11.33s per epoch. 
Although the extra time caused by our multi-branch mechanism is inevitable, our method demonstrates the significance of submanifold for visual perception.
\subsection{Ablation Study}

\begin{table}
\small
  \centering
  \begin{tabular}{ccc}
    \toprule
     Configuration  &  SPDNet  &  MSNet  \\
    \midrule
    \{100,60,36\} & 56.73  & 57.33  \\
    \{100,80,60,36\} & 56.27  & 58.33  \\
    \{100,80,50,25\} & 48.47  & 52.67  \\
    \{100,80,50,25,16\} & 40.07  & 45.60  \\
    \{100,80,50,25,9\} & 28.73  & 36.07  \\
    \bottomrule
    \end{tabular}
  \caption{Performance (\%) on the UCF-sub dataset under different backbone configurations.}
  \label{tab:ablation_config}
\end{table}

To further evaluate the utility of our MSNet, different configurations, like depth and transformation matrices in SPD backbone, are implemented on the UCF-sub dataset, as shown in \tabref{tab:ablation_config}. 
Apart from the expected consistent performance gain brought about by our submanifold block, there is another interesting observation.
The magnitude of improvement varies under different configurations.
To be more specific, in some configurations, like \{100,60,36\}, the improvement sourced from our MSNet is relatively marginal, while in other cases, our approach offers more incremental gain, especially in the case of \{100,80,50,25,9\}, where the SPDNet is highly underfitting.
This indicates that by providing complementary geometrically local information, submanifold is not only beneficial for Riemannian deep learning and could alleviate underfitting.
It is therefore expected that the study of submanifold is worthwhile in the sense of promoting Riemannian deep learning forward.

\section{Conclusion}
In this paper, we successfully identify local mechanisms in Riemannian manifolds and propose a novel multi-scale submanifold block for SPD networks.
Extensive experiments demonstrate the superiority of our approach.
To the best of our knowledge, this work is the first attempt to mine the diverse local geometry in the Riemannian deep network paradigm.
It opens a new direction that mines the information in a high-level semantic submanifold.

However, there are still some issues to be improved. 
For instance, our manual principle for selection is a sub-optimal expedient.
In the future, we will explore other techniques for better selection.
In addition, although local patterns are successfully defined, we rely on approximation in extracting local information.
In the future, we will explore other more intrinsic ways to better deal with submanifolds.

\section*{Acknowledgments}
This work was supported by the National Natural Science Foundation of China (U1836218, 62020106012, 61672265, 62106089), the 111 Project of Ministry of Education of China (B12018), and the Engineering and Physical Sciences Research Council (EPSRC) (EP/N007743/1, MURI/EPSRC/DSTL, EP/R018456/1).

\bibliography{aaai23}
\clearpage
\appendix
\section{Preliminaries}
\subsection{A Brief Review of Differential Geometry}
In this section, we will introduce some basic concepts related to regular submanifolds.
For more detailed discussion, please kindly refer to \cite{loring2011introduction}.

Manifolds are locally Euclidean space, which can be described by coordinate system.
Many ideas in Euclidean space can be generalized into manifolds by locally defined coordinate systems.

\begin{definition}[Coordinate Systems, Charts, Parameterizations] \label{def:Parameterization}
A topological space $\calM$ is locally Euclidean of dimension $n$ if every point in $\calM$ has a neighborhood $U$ such that there is a homeomorphism $\phi$ from $U$ onto an open subset of $\mathbb{R}^{n}$. 
We call the pair $\{ U, \phi: U \rightarrow \mathbb{R}^{n}\}$ as a chart, $U$ as a coordinate neighborhood, the homeomorphism $\phi$ as a coordinate map or coordinate system on $U$, and $\phi^{-1}$ as a parameterization of $U$. 
\end{definition}

$\cinf$-compatibility and atlas are also required to describe the smoothness of manifolds.
Intuitively, an atlas can be viewed as the counterpart on manifolds of standard coordinate system in Euclidean space.

\begin{definition}[$C^{\infty}$-compatible] \label{def:compatible}
Two charts $\{ U, \phi_1: U \rightarrow \mathbb{R}^{n} \},\{ V, \phi_2: V \rightarrow \mathbb{R}^{n} \}$ of a locally Euclidean space are $C^{\infty}$-compatible if the following two composite maps
\begin{equation}
    \begin{aligned}
        \phi_1 \circ \phi_2^{-1} &: \phi_2(U \cap V) \rightarrow \phi_1(U \cap V), \\
        \quad \phi_2 \circ \phi_1^{-1} &: \phi_1(U \cap V) \rightarrow \phi_2(U \cap V)        
    \end{aligned}
\end{equation}
are $\cinf$.
\end{definition}

\begin{definition}[Atlases] \label{def:atlas}
A $C^{\infty}$ atlas or simply an atlas on a locally Euclidean space $\calM$ is a collection $\calA=\{ \{ U_{\alpha}, \phi_{\alpha} \} \}$ of pairwise $\cinf$-compatible charts that cover $\calM$.
\end{definition}

Regular submanifolds are the generalization of subspaces in Euclidean spaces.
Although manifolds usually do not have a standard coordinate system like the Euclidean space, they have atlases.
Hence, regular submanifolds can be defined by atlases.

\begin{definition} [Regular Submanifolds]
A subset $\calS$ of a smooth manifold $\calN$ of dimension $n$ is a regular submanifold of dimension $k$ if for every $p \in \calS$ there is a coordinate neighbourhood $(U, \phi)=\left(U, x^{1}, \ldots, x^{n}\right)$ of $p$ in the maximal atlas of $\calN$ such that $U \cap \calS$ is defined by the vanishing of $n-k$ of the coordinate functions.
\end{definition}

In this paper, we focus on regular submanifolds, since, compared with other kinds of submanifolds, they inherit more properties from manifolds.
From now on, a “submanifold” will always mean a regular manifold. 

Embedding can generate submanifolds.
In some literature, like \cite{lee2013smooth}, this is how regular submanifolds are defined.
Intuitively, embedding is a bijection preserving some associated mathematical properties between domain and its image.
\begin{definition} [Embeddings] \label{def:embedding}
    A $C^{\infty}$ map $f: N \rightarrow M$ is called an embedding if
    \begin{itemize}
        \item it is a immersion and
        \item it is a homeomorphism between $N$ and the image $f(N)$ with the subspace topology.
    \end{itemize}
\end{definition}

\begin{theorem} [Embedded Submanifolds]\label{thm:img_of_embedding}
    If $f: N \rightarrow M$ is an embedding, then its image $f(N)$ is a regular submanifold of $M$.
\end{theorem}

As embeddings can generate submanifolds, the regular submanifold have a more familiar name in machine learning, embedded submanifold.

\subsection{SPD Geometry}
$\mathcal{S}^{d}$ denotes the set of $ d \times d$ real symmetric matrices. 
Then $ S \in \mathcal{S}_{++}^{d}$ iff for all non-zero vector $v \in \mathbb{R}^{d}$, $v^{T} S v>0$. 
As shown in \cite{arsigny2005fast}, $\spd{d}$ forms a Riemannian manifold.
Matrix logarithm $\log (\cdot): \mathcal{S}_{++}^{d} \rightarrow \mathcal{S}^{d}$
is a diffeomorphism \cite{arsigny2005fast}.
This indicates that $\spd{d}$ can be viewed as equivalent to $\sym{d}$ in the perspective of differential manifolds.
This what we apply in the main paper to fuse submanifolds.

\subsection{Direct Sum in Linear Algebra}

In linear algebra, the union of subspaces might not be a linear space.
To generalize the idea of union, there is a sum operation for subspaces.
In particular, there is a direct sum which requires some uniqueness in the construction.
The idea of concatenation in deep learning can be more formally described as direct sum.
\begin{definition}
Supposing $V_1,\cdots,V_k$ are linear subspaces of the linear space $V$, we define direct sum $V_0=V_1 \oplus V_2 \oplus \cdots \oplus V_k$ such that
\begin{equation}
    \forall \alpha \in V_0, \exists ! v_1 \in V_1, \cdots, \exists ! v_k \in V_k, \alpha = v_1 + \cdots + v_k,
\end{equation}
where $+$ is the element addition in linear space $V$.
\end{definition}

\subsection{Covariance Descriptor}
The covariance descriptor, a widely used statistical representation for visual data, encodes correlation of the samples. 
For a given image set $X$, its covariance can be computed as follows,
\begin{equation} \label{eq:cov}
C =\frac{1}{n-1}X H_d X^T,
\end{equation}
where the mean vector of $X$, namely $\mu$, is implicitly computed by the centring matrix  $H_d=I_d-\frac{1}{n}1_d 1_{d}^{T}$.
As studied in \shortcite{pennec2006riemannian}, the covariance matrix lies on an SPD manifold $\mathcal{S}_{++}^{d}$ associated with
\begin{equation} \label{eq:cov_to_SPD}
    S = C + \lambda \operatorname{tr}(C) I_d.
\end{equation} 
In the main paper, we set $\lambda = 1e^{-3}$.

\subsection{Operations on the SPDNet}
By analogy to the conventional fully connected networks, Huang and Van Gool \cite{huang2017riemannian} proposed an SPDNet to perform deep learning that respects the SPD manifold. 
Let  $S_{k-1} \in \mathcal{S}_{++}^{d}$ be the input SPD matrix to the $k^{th}$ layer. We differentiate between the following SPD layers:
\begin{itemize}
\item BiMap layer: This layer is designed to transform the input SPD matrices
to compact and discriminative SPD matrices by $S_{k}=W_{k} S_{k-1} W_{k}^{T}$.
To guarantee the output to be SPD, the transformation matrix $W_{k}$ is constrained to be semi-orthogonal.
\item ReEig layer: This layer serves as an activation function like RELU in a conventional deep network. It is responsible for introducing non-linearity into SPDNet. 
It is defined as follows: 
$S_k = U_{k-1}\max(\epsilon I,\Sigma_{k-1})U_{k-1}^T$, where $\epsilon>0$ is a threshold and $U_{k-1},\Sigma_{k-1}$ define the eigenvalues decomposition, $S_{k-1} = U_{k-1} \Sigma_{k-1}U_{k-1}^T$. 
\item LogEig layer: This layer aims to map the SPD manifold into a Euclidean space. In this way, a Euclidean loss function can be used. 
This layer is defined by matrix logarithm: 
$S_k = U_{k-1}\log(\Sigma_{k-1})U_{k-1}^T$. 
The LogEig layer, together with fully connected layer, softmax and cross entropy, are used to solve classification tasks with SPD representations.
\end{itemize}

\section{Proof for Theorem 1}
\begin{proof}
    Apparently, for arbitrary $\calI$, $(\spd{n})_{\calI} = \spd{k}$ and $(\spd{n})_{\calI}$ is indeed an SPD manifold with a lower dimension.
    The rest is to show that $\spd{k}$ can be embedded into $\spd{n}$.
       
    We first define a map $\phi: \spd{k} \rightarrow \spd{n}$ as the following
    \begin{equation}
        \phi(S') = \exp \left( \left[\begin{array}{cc}
             \log(S') & 0  \\
             0 & 0
        \end{array}\right] \right),
    \end{equation}
    where $S' \in \spd{k}$, and $\exp$ / $\log$ denotes the matrix exponential / logarithm.
    It is easy to verify that $\phi$ is well-defined and is an embedding.
    By \thmref{thm:img_of_embedding}, $\spd{k}$ can be embedded into $\spd{n}$ as a regular submanifold.
\end{proof}

\end{document}